\documentclass{Interspeech2024}

\usepackage{makecell}
\usepackage{bbding}
\usepackage{comment}
\usepackage{multirow}
\usepackage{threeparttable}

\interspeechfinaltrue

\title{PolySpeech: Exploring Unified Multitask Speech Models for Competitiveness with Single-task Models}

\name{Runyan}{Yang}
\name{Huibao}{Yang}
\name{Xiqing}{Zhang}
\name{Tiantian}{Ye}
\name{Ying}{Liu}
\name{Yingying}{Gao}
\name{\\Shilei}{Zhang$^*$}
\name{Chao}{Deng}
\name{Junlan}{Feng}
\address{China Mobile Research Institute, China}
\email{\{yangrunyan,yanghuibao,zhangxiqing,yetiantian,liuyingzn,gaoyingying,\\zhangshilei,dengchao,fengjunlan\}@chinamobile.com}


\begin{document}

\maketitle
 
\begin{abstract}

Recently, there have been attempts to integrate various speech processing tasks into a unified model. However, few previous works directly demonstrated that joint optimization of diverse tasks in multitask speech models has positive influence on the performance of individual tasks. In this paper we present a multitask speech model -- PolySpeech, which supports speech recognition, speech synthesis, and two speech classification tasks. PolySpeech takes multi-modal language model as its core structure and uses semantic representations as speech inputs. We introduce semantic speech embedding tokenization and speech reconstruction methods to PolySpeech, enabling efficient generation of high-quality speech for any given speaker. PolySpeech shows competitiveness across various tasks compared to single-task models. In our experiments, multitask optimization achieves performance comparable to single-task optimization and is especially beneficial for specific tasks.

\end{abstract}

\ifinterspeechfinal
    \renewcommand{\thefootnote}{}
    \footnotetext{*Corresponding author.}
    \renewcommand{\thefootnote}{\arabic{footnote}}
\fi

\vspace{0.1in}
\section{Introduction}

Recently researchers have proposed model architectures and training techniques to aggregate various speech processing tasks into one unified framework. Performance of these frameworks demonstrate the capability of deep learning models like Transformer\cite{NIPS2017_7181,Zhou2018} to learn and express the knowledge in various modalities required for diverse speech tasks.

Classic encoder-decoder architecture has been explored to model multiple speech tasks. For example, 
Whisper\cite{radford2023robust} is capable of performing speech processing tasks such as speech recognition, speech translation, and language identification, utilizing large-scale multilingual supervised speech as training data for an encoder-decoder Transformer.
There are also recent attempts to aggregate multiple speech tasks in simpler decoder-only language model (LM) structures instead of encoder-decoder. VioLA\cite{wang2023viola}, AudioPaLM\cite{rubenstein2023audiopalm}, LauraGPT\cite{chen2023lauragpt}, SpeechGPT\cite{zhang2023speechgpt}, etc. support speech transcription and generation tasks by modeling both speech and text representations with Transformer LMs. Speech representations used by these models can be categorised into two types -- acoustic ones and semantic ones. Acoustic speech representations are generated from speech waveform through speech codec methods, e.g. EnCodec\cite{defossez2022high} and SoundStream\cite{zeghidour2021soundstream}. Semantic representations are usually extracted from self-supervised learning models such as wav2vec\cite{schneider2019wav2vec,baevski2020wav2vec,chung2021w2v} and HuBERT\cite{hsu2021hubert}.

Previous literature\cite{wang2023viola,rubenstein2023audiopalm,chen2023lauragpt} has analyzed the performance differences between the proposed multitask speech models and existing single-task models, but few studies compare these models under fair enough experimental conditions, such as same supervised training data.
The aim of this paper is to demonstrate that multitask models are in deed competitive in performance on various tasks compared to single-task models. 
We intend to find out whether joint optimization of various speech tasks within a single auto-regressive Transformer decoder framework improves performance comparing to single-task optimization.
We are also curious whether the multitask model benefits more from semantic speech representations or from acoustic ones.

In this paper, we propose a multitask speech model framework, which we call PolySpeech. PolySpeech's core structure is a multi-modal decoder-only Transformer LM, which autoregressively predicts speech or text tokens. We integrate tasks of speech recognition (ASR), speech synthesis (TTS), spoken language identification (LID), and gender identification (GID) in PolySpeech. These tasks, covering three main types of speech tasks -- transcription, generation, and classification, are jointly optimized in a supervised manner. We prefer to using semantic speech representations rather than using acoustic ones in PolySpeech for better performance. PolySpeech is highly flexible and can be further extended to other speech tasks. 



Discretization of speech representations is crucial for autoregressive prediction models to generate speech. In preliminary experiments, we have found that the k-means discretization method used in AudioLM\cite{borsos2023audiolm} and Spear-TTS\cite{kharitonov2023speak} would lead to loss of acoustic information. We tried to resynthesize HuBERT k-means tokens of Mandarin Chinese speech with a HiFi-GAN\cite{kong2020hifi} vocoder and observed tone inaccuracies in synthesized speech, which are unacceptable for Mandarin. 
For better speech generation performance, we use a codec method for semantic embedding of speech in PolySpeech, which preserves more complete acoustic information in discrete tokens than the k-means method.
Focusing on producing high quality speech while controlling acoustic conditions, we also design a semantic speech token decoder that generates speech waveform given discrete semantic tokens and a speech prompt from any speaker.


We summarize the contribution of this paper as follows: 
\begin{enumerate} 
\item We propose a multitask speech model based on multi-modal LM and semantic speech representations, which is competitive across various tasks compared to single-task models.
\item 
We introduce a high-fidelity speech codec method and a semantic speech token decoder to the multitask speech model, enabling efficient speech generation for any given speaker.
\item We conduct meaningful experiments, demonstrating that multitask optimization achieves performance on par with single-task optimization and is beneficial for certain tasks.
\end{enumerate} 


\section{PolySpeech}
\label{sec:method}

In this section we describe PolySpeech, our proposed multitask speech model framework. A diagram of PolySpeech is illustrated in Figure \ref{fig:polysimple}.

\subsection{Model architecture}

\begin{figure}[t]
  \centering
  \includegraphics[width=0.9\linewidth]{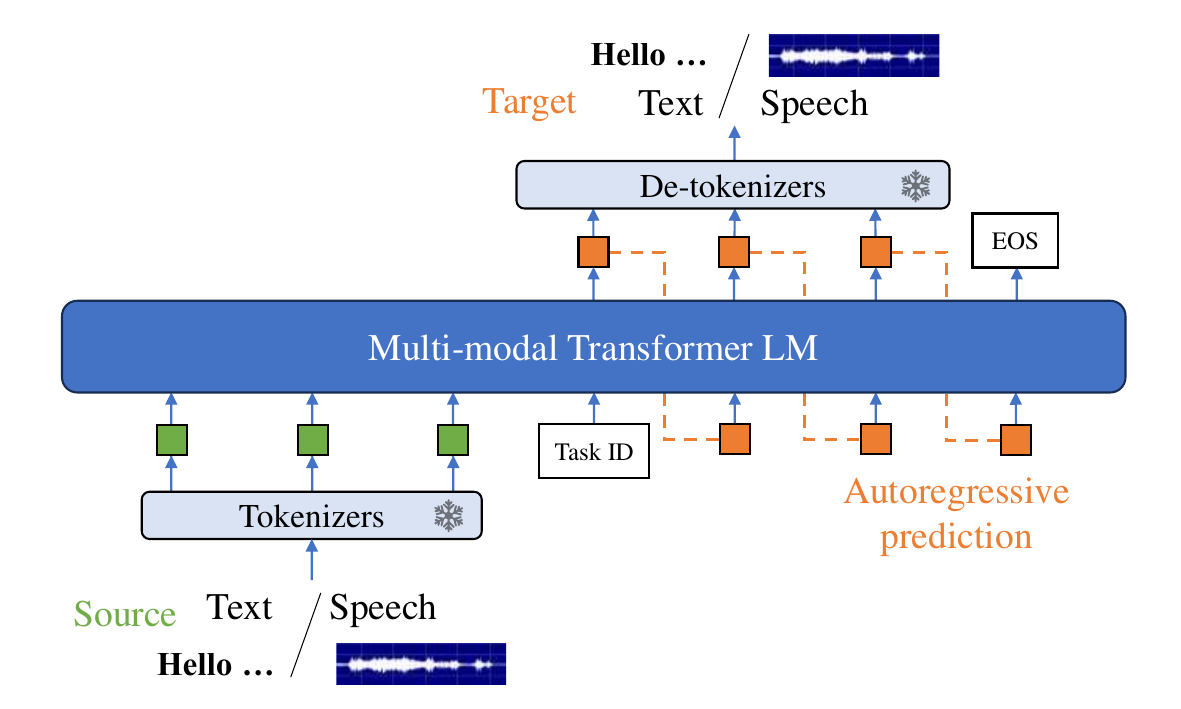}
  \vspace{-0.1in}
  \caption{A brief diagram of the PolySpeech framework.}
  \label{fig:polysimple}
  \vspace{-0.15in}
\end{figure}

As shown in Figure \ref{fig:polysimple}, PolySpeech consists of three basic parts -- a decoder-only Transformer-based multi-modal LM, tokenizers for inputs, and de-tokenizers for outputs. 

The multi-modal Transformer LM is the core structure of PolySpeech. Given discrete tokens of a speech or text sequence, which is the source sequence of a speech task, the LM predicts the task's target token sequence autoregressively. The target sequence is also in speech or text modality, depending on task. The LM embeds input token sequences from each modality into a continuous vector space using modal-specific token embedding layers, and uses task-specific linear layers to generate output target logits. When we take multi-dimensional speech tokens as model input, such as 8-dimensional EnCodec\cite{defossez2022high} tokens, we use different embedding layers to embed tokens from each dimension and average each embedding layer's output. 
We also attempt to directly use continuous embedding vectors instead of using discrete tokens as the LM's input source speech sequence. In this case, a linear layer is used to map the source embedding vectors into the LM's input vector space.

Besides the LM, PolySpeech also contains input tokenizer and output de-tokenizer modules for each modality. These modules are prepared before the training of the LM and not optimized together with the LM. For text modality, the tokenizer maps original text into sequences of text tokens, while the de-tokenizer maps the LM's output tokens to text. For speech modality, the tokenizer can be any type of audio encoder that converts speech waveform into discrete speech tokens or continuous speech embeddings, such as EnCodec or HuBERT, while the de-tokenizer is a speech token decoder that constructs waveform given the LM's output tokens.

\subsection{Model input and output}

Source and target sequence of a task are concatenated as the input of the multi-modal LM. A task-specific ``task ID'' token in inserted between the source and target sequences, enabling PolySpeech to distinguish different tasks. PolySpeech handles task ID tokens with a special token embedding layer.

During training, ground-truth target sequence is shifted left and used as training targets. Cross entropy loss is computed between training targets and logits predicted by the LM. In addition, an upper triangular attention mask is applied to the target sequence to ensure causality.

During inference, first the source sequence and task ID are feed into the LM, and then the model predict target sequence in an autoregressive manner. Top-K sampling and beam search methods are used for inference with speech and text modality outputs, respectively. The inference process ends when the LM outputs an ``end of sentence (EOS)'' token.

In the following sections, we will describe the input and output sequences as well as speech tokenizers and de-tokenizers for each task. Brief information can also be found in Table \ref{tab:task_info}.

\begin{table}[thbp]
  \caption{Source and target sequences of each task.}
  \label{tab:task_info}
  \vspace{-0.05in}
  \centering
  \footnotesize
  \begin{tabular}{ c c c }
    \toprule
    \textbf{Task} & \textbf{Source sequence} & \textbf{Target sequence} \\
    \midrule
    ASR & speech tokens/embeddings & text tokens \\
    TTS & text tokens & speech tokens \\
    LID & speech tokens/embeddings & language tag \\
    GID & speech tokens/embeddings & gender tag \\
    \bottomrule
  \end{tabular}
  \vspace{-0.15in}
\end{table}

\subsubsection{ASR}

Source and target sequences of the ASR task are in speech and text modalities, respectively. Either semantic-based speech token/embedding or acoustic speech token/embedding can be used as source sequence. 
We prefer to semantic speech token/embeddings rather than acoustic ones for better performance.
Text transcript of the speech utterance is converted to a token sequence using lexicons.


\subsubsection{TTS}

PolySpeech is able to synthesize speech for any in-domain or out-of-domain speaker.
The source and target sequences of TTS are exactly inverse to those of ASR.
In TTS task, the LM predicts tokens' logits for target speech, which means that we must represent speech in a discrete form. The discretization method we take will be presented later in Section \ref{sec:repcodec}.

To accomplish TTS for any given speaker, PolySpeech receives a speech prompt for speaker information. During inference, we first sequentially input the following tokens into the LM, and then predict target speech tokens: the speech prompt's transcript text tokens, TTS source text tokens, TTS task ID, and the speech prompt's speech tokens. The speech prompt is also fed into the speech de-tokenizer, which will be introduced in Section \ref{sec:detokenizer}.

\subsubsection{Classification tasks (LID and GID)}

The speech classification tasks share the same source sequence with ASR. The target sequence is a classification tag, which can also be considered as a special text token sequence of length 1.

\subsection{Speech tokenizer and de-tokenizer for TTS}


\subsubsection{Semantic speech embedding tokenization}
\label{sec:repcodec}


In order to efficiently descritize semantic speech embeddings for TTS task, we apply a codec method, which we call Semantic Speech Embedding Tokenization (SSET). In SSET, a model that shares a similar structure as AudioDec\cite{wu2023audiodec} is used, which consists of a convolutional encoder-decoder network and a residual vector quantizer (RVQ). During training, both input and output are continuous semantic speech representation, and the objective function is L2 loss. In PolySpeech we use 
the output of the first RVQ layer as discrete speech tokens. The SSET we proposed allows for a high-fidelity preservation of acoustic information.  

\begin{figure}[t]
  \centering
  \includegraphics[width=0.95\linewidth]{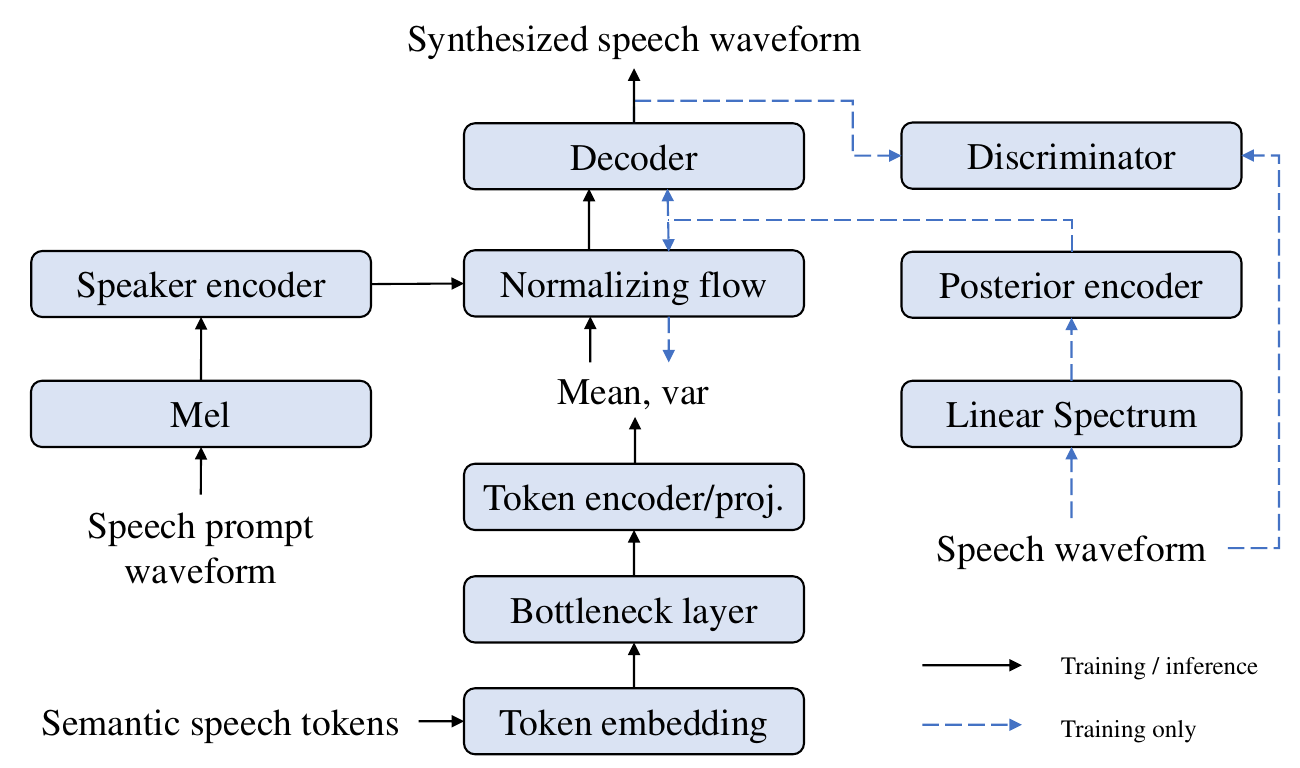}
  \vspace{-0.05in}
  \caption{Diagram of the semantic speech token decoder.}
  \label{fig:codec_decoder}
  \vspace{-0.15in}
\end{figure}

\subsubsection{Speech waveform generation}
\label{sec:detokenizer}
Drawing inspiration from VITS\cite{kim2021conditional}, we design a non-autoregressive semantic token decoder to generate speech waveforms using semantic speech tokens. The decoder's model architecture is illustrated in Figure \ref{fig:codec_decoder}. We have made several essential modifications on the basis of the VITS architecture for fast and efficient waveform generation. First, since we use semantic speech tokens rather than text, which are already aligned to the speech waveform, we omit the alignment estimation methods used in VITS. Second, speech prompts are used to control acoustic features, enabling TTS for any given speaker. We use a speaker encoder consisting of LSTM, linear layer, and ReLU activation function to extract the speaker embedding of the speech prompt as an input to the decoder. What's more, we replace the HiFi-GAN vocoder with VOCOS\cite{siuzdak2023vocos}, which generates high-fidelity speech and accelerates inference speed.

\subsection{Parameter initialization and multitask optimization}
\label{sec:lm}

Before PolySpeech training, we initialize the multi-modal LM's parameters as well as the text token embedding layer parameters using a well-trained text-modality-only Transformer LM. 

For each batch of the PolySpeech optimization process, we randomly select a task and calculate gradient using training data from this task. The probability of each task being selected depends on the proportion of this task's training data volume to the total training data volume of all tasks. The gradients of several batches are accumulated for a single update of the model.



\section{Experimental settings}
\label{sec:setting}

\subsection{Datasets}
Standard Mandarin Chinese speech corpora are used for ASR and TTS tasks. Aishell-1\cite{bu2017aishell}, Aishell-2\cite{du2018aishell}, WenetSpeech\cite{zhang2022wenetspeech}, KeSpeech\cite{tang2021kespeech}, and our internal Mandarin dialog corpora are used together as the training set. The total amount of speech used for ASR and TTS training is 26k hours. We use development subsets of the four open-source corpora for validation and use Aishell-1 test for evaluation. In ASR experiments, we also train models solely on Aishell-1 to ensure a fairer comparison with baseline and enhance experiment efficiency.

Training data of LID and GID tasks consists of speech in four Chinese dialects -- Standard Mandarin, Cantonese, Hokkien, and Southwestern (SW) Mandarin -- as well as English. Information of these corpora is illustrated in Table \ref{tab:class_trainset}. We resample the training data to maintain balance in data volume for each classification category. For LID, we randomly sample 500 hours of speech from the corpora of each language. For GID, we randomly sample 500 hours of speech for each gender (male and female) from the entire corpora. Validation sets are sampled from the same sources of corresponding training sets. 
We use out-of-distribution data to evaluate classification performance of models. LID test set information is listed in Table \ref{tab:class_trainset}. 
For GID, We use ASR-SCKwsptSC
\footnote{https://magichub.com/datasets/mandarin-chinese-scripted-speech-corpus-keyword-spotting/}
\footnote{https://magichub.com/datasets/mandarin-chinese-scripted-speech-corpus-keyword-spotting-2/} 
as test set, which contains 9613 and 5962 utterances from female and male speakers, respectively. Utterances shorter than 1.6s are excluded from classification training, validation, and test sets.

WuDaoCorpora Text\cite{yuan2021wudaocorpora} is used as the training data for the text-only LM mentioned in Section \ref{sec:lm}.

\begin{table}[thbp]
  \caption{LID and GID training corpora (before sampling).}
  \label{tab:class_trainset}
  \vspace{-0.05in}
  \centering
  \footnotesize
  \begin{tabular}{ c c c c }
    \toprule
    \textbf{Dataset} & \textbf{Language(dialect)} & \textbf{Hours} & \textbf{Style} \\ 
    \midrule
    Aishell-2 & Standard Mandarin & 1000 & reading \\
    internal & Standard Mandarin & 1787 & dialog \\
    internal & Cantonese & 1250 & dialog \\
    internal & Hokkien & 451 & dialog \\ 
    internal & SW Mandarin & 1500 & dialog \\
    LibriSpeech\cite{panayotov2015librispeech} & English & 961 & reading \\

    \bottomrule
  \end{tabular}
  \vspace{-0.15in}
  
\end{table}

\begin{table}[thbp]
  \caption{LID test set information.}
  \label{tab:lid_testset}
  \vspace{-0.05in}
  \centering
  \footnotesize
  \begin{threeparttable}
  \begin{tabular}{ c c c }
    \toprule
    \textbf{Dataset Name} & \textbf{Language(dialect)} & \textbf{\# utts} \\ 
    \midrule
    FLEURS\cite{fleurs2022arxiv} yue & Cantonese & 819 \\
    Common Voice\cite{ardila2019common} nan-tw & Hokkien & 1203\tnote{*} \\
    ASR-SCWuhDiaDuSC\footnotemark & SW Mandarin & 1000\tnote{*} \\
    FLEURS cmn & Standard Mandarin & 944 \\
    FLEURS en & English & 647 \\  
    
    \bottomrule
  \end{tabular}
  \begin{tablenotes}
    \footnotesize
    \item[*] Randomly sampled from the original dataset.
  \end{tablenotes}
  \end{threeparttable}
  \vspace{-0.15in}
\end{table}

\footnotetext[3]{https://magichub.com/datasets/wuhan-dialect-scripted-speech-corpus-daily-use-sentence/}

\subsection{Model and training settings}

The multi-modal LM in PolySpeech is a 12-layer Transformer. 
Each layer includes 12 attention heads, 768 hidden units, and 3,072 feed-forward units.
Chinese characters are used as ASR output and TTS input, along with Latin letters and special labels. 
The source sequences of ASR, LID, and GID tasks are 768-dimensional embedding vectors extracted from the output of the ninth encoder layer of chinese-hubert-base\footnote{https://huggingface.co/TencentGameMate/chinese-hubert-base}. Polyspeech contains a total of 160M trainable parameters.
SSET described in Section \ref{sec:repcodec} are applied to the outputs of the ninth layer of chinese-hubert-base to generate TTS targets. We use WenetSpeech to train the SSET model and configure the number of clusters to be 2048. The semantic token decoder is trained using Aishell-1 and WenetSpeech corpora.

PolySpeech is trained with 16 NVIDIA V100 32GB GPUs, with a total batch size of 20 minutes of speech. Gradients are accumulated for 4 steps before a weight update is performed. Adam algorithm\cite{adam} with warm-up of 2k steps is used. We select the model with the lowest validation loss for evaluation.

\subsection{Evaluation Metrics}
We use character error rate (CER) to evaluate Chinese ASR performance.
For TTS, we use two objective metrics. The first one is CER of the synthesized speech, which is evaluated with Whisper Large-v3 model. The second one is speaker encoder cosine similarity (SECS)\cite{casanova2021sc} calculated between synthesized speech and speech prompt using a speaker varification model\footnote{https://huggingface.co/microsoft/wavlm-base-plus-sv}. CER and SECS measure pronunciation accuracy and timbre similarity for the synthesized speech, respectively. Classification accuracy and Macro F1 score are used to evaluate performance of two speech classification tasks.

\section{Results and analysis}
\label{sec:result}

\subsection{ASR task}

ASR results are exhibited in Table \ref{tab:asr_result}.
We use result of a typical encoder-decoder transformer ASR model\footnote{https://github.com/espnet/espnet/blob/master/egs/aishell/asr1} trained with Espnet\cite{watanabe2018espnet} as a single-task baseline to compare PolySpeech with. We perform ASR decoding using beam search algorithm with beam size 5. No external LMs are used in decoding. Results of our final PolySpeech model are presented in Table \ref{tab:asr_result} and other result tables with \textbf{bold font}.

It is reported from the results of models trained on Aishell-1 that multiple-task-trained models show better performance than single-task-trained models.
With the training data increasing to 26k hours, multitask training no longer exhibits a significant improvement in ASR CER, remaining performance roughly comparable to single-task training.

From the comparison of speech tokenizers, we find that the model employing continuous HuBERT embedding as input speech representation (``HuBERT'' in the table) yields the best ASR performance. HuBERT tokens generated by SSET perform not as good as continuous embeddings, but they are obviously superior to HuBERT k-means tokens. Semantic HuBERT tokens outperform acoustic EnCodec tokens. Results also indicate that text-only LM initialization improves performance of the ASR task, which involves text generation procedure.

\begin{table}[thbp]
  \caption{PolySpeech ASR results on Aishell-1 test.}
  \label{tab:asr_result}
  \vspace{-0.05in}
  \centering
  \footnotesize
  \begin{tabular}{ c c c c c }
    \toprule
    \makecell[c]{\textbf{ASR} \\ \textbf{trainset}} & \makecell[c]{\textbf{Training} \\ \textbf{task(s)}} & \makecell[c]{\textbf{ASR speech} \\ \textbf{tokenizer}} & \makecell[c]{\textbf{LM} \\ \textbf{init.}} & \makecell[c]{\textbf{CER} \\ \textbf{\%}} \\

    \midrule
    Aishell-1 & \multicolumn{3}{c}{ESPNet Transformer (baseline)} & 7.4 \\
    
    \midrule
    Aishell-1 & ASR & EnCodec & \XSolidBrush & 18.1 \\
    Aishell-1 & ASR & HuBERT k-means & \XSolidBrush & 14.0 \\
    Aishell-1 & ASR & HuBERT codec & \XSolidBrush & 8.1 \\
    Aishell-1 & ASR & HuBERT & \XSolidBrush & 7.1 \\
    Aishell-1 & ASR+TTS & HuBERT & \XSolidBrush & 6.7 \\
    Aishell-1 & ASR+TTS & HuBERT & \Checkmark & 6.0 \\
    26k hrs & ASR & HuBERT & \Checkmark & 2.7 \\
    26k hrs & ASR+TTS & HuBERT & \Checkmark & 2.7 \\
    \textbf{26k hrs} & \textbf{4 tasks} & \textbf{HuBERT} & \Checkmark & \textbf{2.0} \\
    
    \bottomrule
  \end{tabular}
  \vspace{-0.2in}
  
\end{table}

\subsection{TTS task}

In our preliminary experiments, we found that PolySpeech models trained solely on Aishell-1 were not able to provide stable enough TTS performance, especially when generating long speech.
When we expand the training data to 26k hours, the TTS performance get greatly improved. In Table \ref{tab:tts_result}, we present PolySpeech TTS performance evaluated on Aishell-1 test subset. Since there are no other publicly available results of models trained on the same data set, we only show the performance of PolySpeech trained on different speech tasks, using original Aishell-1 test speech as topline. We set k to 5 for the top-k sampling in the TTS experiments. 

Results indicate that the TTS performance of PolySpeech models trained on multiple tasks either surpasses or closely matches that of models trained solely on TTS. The CER of 4-tasks PolySpeech model increases by only absolute 0.3\% comparing to the original speech and the SECS is larger than 0.9, which indicates that the synthesized speech is correct in pronunciation and closely resembles the speech prompt in timbre.

\begin{table}[thbp]
  \caption{PolySpeech TTS results on Aishell-1 test subset.}
  \label{tab:tts_result}
  \vspace{-0.05in}
  \centering
  \footnotesize
  \begin{tabular}{ c c c c c c }
    \toprule
    \textbf{Training task(s)} & \textbf{TTS trainset} & \textbf{CER \%} & \textbf{SECS} \\
    \midrule
    \multicolumn{2}{c}{Original speech} & 4.6 & 0.94 \\
    \midrule
    TTS & 26k hrs & 5.1 & 0.91 \\
    ASR+TTS & 26k hrs & 5.4 & 0.91 \\
    \textbf{4 tasks} & \textbf{26k hrs} & \textbf{4.9} & \textbf{0.91} \\
    \bottomrule
  \end{tabular}
  \vspace{-0.2in}
  
\end{table}

\subsection{Classification tasks (LID and GID)}

LID and GID results on the test sets are shown in Table \ref{tab:class_result}. 
As a single-task baseline model, we fully finetuned Whisper Small on the same LID and GID training set as PolySpeech. 

The finetuned Whisper baseline and PolySpeech models trained with EnCodec tokens work well on the validation set, with accuracy exceeding 99\% (not listed in the table), but perform poorly on out-of-domain test sets, especially for the LID task. 
Since each language's LID training speech come from different data sources, models trained with EnCodec acoustic tokens are likely to overfit to the acoustic conditions of each language's training set.
In contrast, models using HuBERT semantic embeddings demonstrate better performance on the test set, showing strong generalization ability.

Multitask training brings significant performance to models using HuBERT on LID task. This is because the model further maintains semantic discrimination abilities through the ASR task, which is beneficial to LID. And since we use Standard Mandarin speech in ASR training,
Standard Mandarin shows the largest LID accuracy improvement (from 81.3\% to 91.8\%) among the 5 languages and dialects. In GID task, both single-task and multitask models perform well on the test set.

\begin{table}[thbp]
  \caption{PolySpeech LID and GID results on test sets.}
  \label{tab:class_result}
  \vspace{-0.05in}
  \centering
  \footnotesize
  \begin{tabular}{ c c c c c c}
    \toprule
    \multirow{2}{*}{\makecell[c]{\textbf{Training} \\ \textbf{task(s)}}} & \multirow{2}{*}{\makecell[c]{\textbf{Speech} \\ \textbf{tokenizer}}} & \multicolumn{2}{c}{\textbf{Accuracy \%}} & \multicolumn{2}{c}{\textbf{Macro F1 \%}} \\
    & & \textbf{LID} & \textbf{GID} & \textbf{LID} & \textbf{GID} \\
    \midrule
    \multicolumn{2}{c}{Ft. Whisper (baseline)} & 44.4 & 97.7 & 38.6 & 97.5 \\
    \midrule
    LID & EnCodec & 43.4 & - & 38.1 & - \\
    LID & HuBERT & 90.9 & - & 91.1 & - \\
    GID & EnCodec & - & 95.9 & - & 95.6 \\
    GID & HuBERT & - & 99.5 & - & 99.5 \\
    4 tasks & EnCodec & 44.2 & 96.1 & 38.6 & 95.8 \\
    \textbf{4 tasks} & \textbf{HuBERT} & \textbf{94.4} & \textbf{99.0} & \textbf{94.8} & \textbf{98.9} \\
    \bottomrule
  \end{tabular}
  \vspace{-0.2in}
  
\end{table}

\section{Conclusion}
\label{sec:conclusion}
In this paper we present PolySpeech, a multitask speech model exhibiting competitiveness across diverse tasks when compared to single-task models. We also introduce a high-fidelity speech codec method along with an efficient semantic speech token decoder for efficient speech generation. Through experiments, we validate that multitask optimization achieves performance comparable to single-task optimization and offers advantages for certain tasks. In the future, we plan to integrate more speech tasks into PolySpeech, such as speech translation, multilingual ASR, and speech enhancement. We will also expand the model scale of PolySpeech to achieve better performance.

\bibliographystyle{IEEEtran}
\bibliography{mybib}

\end{document}